\documentclass[letterpaper, 10 pt, conference]{ieeeconf}  

\IEEEoverridecommandlockouts                              
\usepackage[utf8]{inputenc}
\usepackage[T1]{fontenc}
\usepackage{float}
\usepackage{graphicx}
\usepackage{amsmath}
\usepackage{listings}
\usepackage{algpseudocode}
\usepackage{algorithm}
\usepackage{xcolor}
\graphicspath{ {./images/} }

\newcommand*{\email}[1]{\texttt{#1}}

\title{\LARGE \bf
Neighbor-Based Optimized Logistic Regression Machine Learning Model For Electric Vehicle Occupancy Detection
}
\author{Sayan Shaw, Keaton Chia, Jan Kleissl\\
\textit{University of California San Diego}\\
\email{\{sashaw, kwchia, jkleissl\}@ucsd.edu}
}

\begin{document}

\maketitle
\thispagestyle{empty}
\pagestyle{empty}

\renewcommand\keywords{Keywords}
\begin{abstract}

This paper presents an optimized logistic regression machine learning model that predicts the occupancy of an Electric Vehicle (EV) charging station given the occupancy of neighboring stations. The model was optimized for the time of day. Trained on data from 57 EV charging stations around the University of California San Diego campus, the model achieved an 88.43\% average accuracy and 92.23\% maximum accuracy in predicting occupancy, outperforming a persistence model benchmark.

\end{abstract}

\section{INTRODUCTION}

The intersection of machine learning (ML) and occupancy detection currently involves artificial Recurrent Neural Network (RNN) models known as Long Short-Term Memory (LSTM) models with layers of several neurons, specializing in processing sequences of data, and modeling long term dependencies [1]. They can be used to predict the occupancy of rooms or usage of appliances in the future given their occupancy or usage in the past (i.e. given $a_1$ predict $a_2$). LSTM models have become the industry standard for making such kinds of predictions due to the fact that their ability to train on sequences of recurrent data results in accurate forecasts for large time-series data [2]. LSTM models have been used in the past to predict vehicle and human occupancy using time-series data. For instance, Gutmann et al. developed a LSTM-based model for predicting truck occupancy using historic data with 50 truck parking spaces and achieved an average root mean square error metric of 3.15 trucks [3]. Kim and Tangrand et al. both describe methods to leverage LSTM models to predict human occupancy using data from sensors measuring environmental variables such as $CO_2$, humidity, and temperature [4]-[5]. However, predicting occupancy of Electric Vehicle (EV) charging stations specifically is a relatively unexplored area. Recently, a LSTM-based model shared by Ma et al. in 2021 predicted EV charging station occupancy based on historic data from 57 stations in the city of Dundee, United Kingdom, achieving 99.99\% and 81.87\% accuracy for predictions made 10 minutes and 1 hour ahead, respectively [6]. Note that all of the models mentioned thus far required historic data for predictions.

The motivation of the work presented in this paper was to build a new, streamlined model that uses a logistic regression classifier and performs occupancy prediction of an EV charging station using the real-time occupancy of neighboring EV charging stations (i.e. given $a, b, c,$ predict $d$), optimized for the time of day in order to make predictions with higher accuracy than the traditional LSTM-based models. We built a neighbor-based optimized logistic regression (NOLR) model using occupancy data from 57 different EV charging stations around the University of California San Diego campus, and achieved an 88.43\% average accuracy and 92.23\% maximum accuracy in predicting occupancy of a station based on the occupancy of its neighboring stations. The novelty in this model thereby lies not only in the fact that it uses a new neighbor-based approach to predict occupancy, but also that it does so with comparable accuracy to LSTM models while only using a single layer model, rather than a multilayered LSTM model.

The remainder of this paper is organized as follows: Section II describes the methods and the engineering of the model. In Section III, we analyze the results we achieve over various ranges of data and compare them to a benchmark model. We conclude this paper in Section IV, where we discuss the implications of our results.

\section{METHODS}

\subsection{Goal}

We first formalize our problem statement in order to ensure accurate parameters, data, and prediction definitions. Our goal was to predict the occupancy of an EV charging station given the occupancy of neighboring stations in a specific time range. A formal definition of our problem statement can thus be worded as follows: Given the binary occupancy data of $n$ different charging stations labeled $a, b, c,\ldots$, with 0 representing unoccupied and 1 representing occupied, for every hour in a specific time range $[1, m]$, predict the occupancy of another charging station $s \not\in {a, b, c,\ldots}$ for each hour in $[1, m]$. 

\subsection{Model Selection}

There are two quintessential types of supervised machine learning problems -- classification and regression -- each with a variety of models specializing in different areas within them. The former deals with discrete output predictions, and the latter requires continuous output predictions. Given the fact that our model needed to deal with binary data, where both input and output values were discrete binary digits signifying occupancy, we needed a classification model. Our data was not only discrete, but also binary, therefore this motivated choosing a logistic regression model, since its sigmoid function acts as a binary classifier.

\subsection{Data Vectorization}

As with any logistic regression model, we needed vectorized input and output data for training and testing our model. Let $x_1, x_2, x_3,\ldots, x_m$ be $n$-length vectors, each with $n$ binary values, which represent the binary occupancy of a charging station at a unique time. These $m$ vectors form a $n \times m$ dimensional input matrix $X = [x_i]$ for $i = 1$ to $i = m$. Similarly, we had a $m$-length output vector $y$ consisting of the occupancy values of charging station $s$ at each hour from $1$ to $m$. Table 1 shows a visualization of this entire process.

\begin{table}[H]
\tiny
\caption{Data Vectorization}
\label{data_vectorization}
\begin{center}
\begin{tabular}{|c|cccccc|cc|}
\hline
\textbf{Time} & \multicolumn{6}{c|}{\textbf{Raw Binary Occupancy Data}} & \multicolumn{2}{c|}{\textbf{Vectorized Model Data}} \\ \hline
\multicolumn{1}{|l|}{} & \multicolumn{1}{c|}{\textit{\textbf{a}}} & \multicolumn{1}{c|}{\textit{\textbf{b}}} & \multicolumn{1}{c|}{\textit{\textbf{c}}} & \multicolumn{1}{c|}{\textbf{. . .}} & \multicolumn{1}{c|}{\textit{\textbf{n}}} & \textit{\textbf{s}} & \multicolumn{1}{c|}{\textbf{$x_i$ from i = 1 to  i = m}} & \textit{\textbf{y}} \\ \hline
\textit{\textbf{1}} & \multicolumn{1}{c|}{1} & \multicolumn{1}{c|}{0} & \multicolumn{1}{c|}{1} & \multicolumn{1}{c|}{. . .} & \multicolumn{1}{c|}{1} & 1 & \multicolumn{1}{c|}{{[}1, 0, 1, . . . , 1{]}} & 1 \\ \hline
\textit{\textbf{2}} & \multicolumn{1}{c|}{0} & \multicolumn{1}{c|}{1} & \multicolumn{1}{c|}{0} & \multicolumn{1}{c|}{. . .} & \multicolumn{1}{c|}{1} & 1 & \multicolumn{1}{c|}{{[}0, 1, 0, . . . , 1{]}} & 1 \\ \hline
\textit{\textbf{3}} & \multicolumn{1}{c|}{0} & \multicolumn{1}{c|}{0} & \multicolumn{1}{c|}{0} & \multicolumn{1}{c|}{. . .} & \multicolumn{1}{c|}{0} & 0 & \multicolumn{1}{c|}{{[}0, 0, 0, . . . , 0{]}} & 0 \\ \hline
\textit{\textbf{. . .}} & \multicolumn{1}{c|}{. . .} & \multicolumn{1}{c|}{. . .} & \multicolumn{1}{c|}{. . .} & \multicolumn{1}{c|}{. . .} & \multicolumn{1}{c|}{. . .} & . . . & \multicolumn{1}{c|}{. . .} & . . . \\ \hline
\textit{\textbf{m}} & \multicolumn{1}{c|}{1} & \multicolumn{1}{c|}{1} & \multicolumn{1}{c|}{1} & \multicolumn{1}{c|}{. . .} & \multicolumn{1}{c|}{1} & 1 & \multicolumn{1}{c|}{{[}1, 1, 1, . . . , 1{]}} & 1 \\ \hline
\end{tabular}
\end{center}
\end{table}

\subsection{Logistic Regression Model}

\begin{figure}[thpb]
  \centering
  \framebox{
  \includegraphics[scale=0.24]{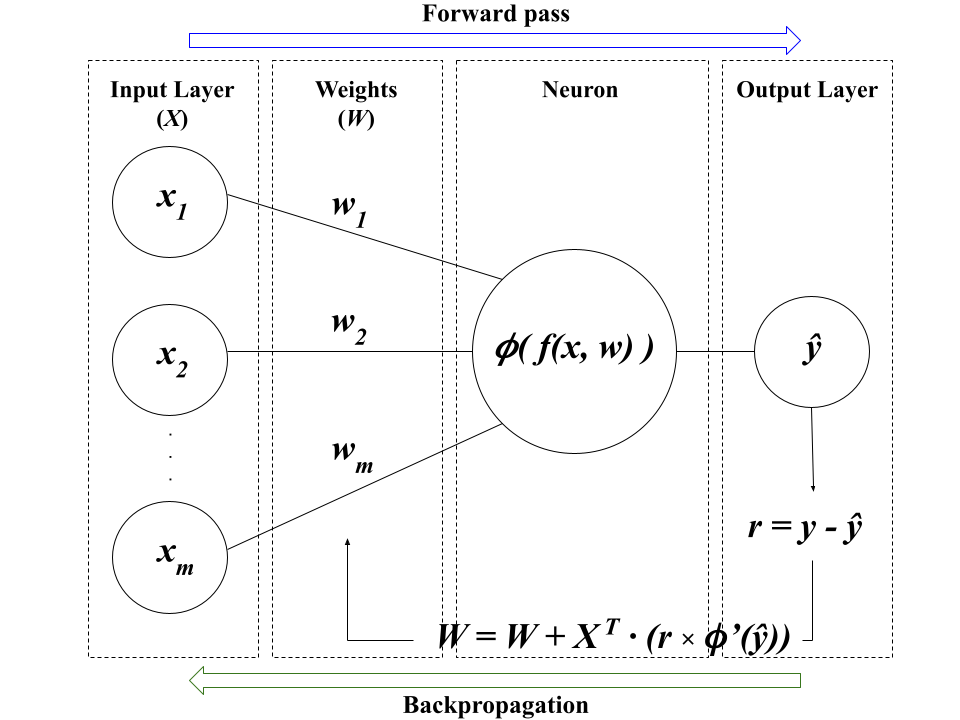}}
  \caption{Visualization of Logistic Regression Model}
  \label{model}
\end{figure}

This subsection explains the functionality of a traditional logistic regression ML model in our use case. It can be visualized from start to finish in Fig.~\ref{model}.

A logistic regression ML model is a single-layer model, with simply one neuron, also known as a single-layer perceptron model. This aligns with our goal of devising a streamlined model. We shall proceed by providing a mathematical analysis of this model that describes how it can be used to predict the binary occupancy of a charging station given the binary occupancy of neighboring charging stations.

The first step in our model pipeline is evaluating a net input function $f(x, w)$ that is the dot product of the $m$ different $n$-length vectors $x_1$ through $x_m$ with weights $w_1$ through $w_m$, which also form a weight matrix $W = [w_i]$ for $i = 1$ to $i = m$. These weights are random floating point values between 0 and 1, and they will be repeatedly changed afterwards for improved accuracy using a process called backward propagation of errors, or backpropagation. This net input function is defined by equation (1).
\begin{equation}
f(x, w) = \sum_{i=1}^{m} x_i w_i = x_1 w_1 + x_2 w_2 + \ldots + x_m w_m
\end{equation}

\begin{figure}[thpb]
  \centering
  \includegraphics[scale=0.68]{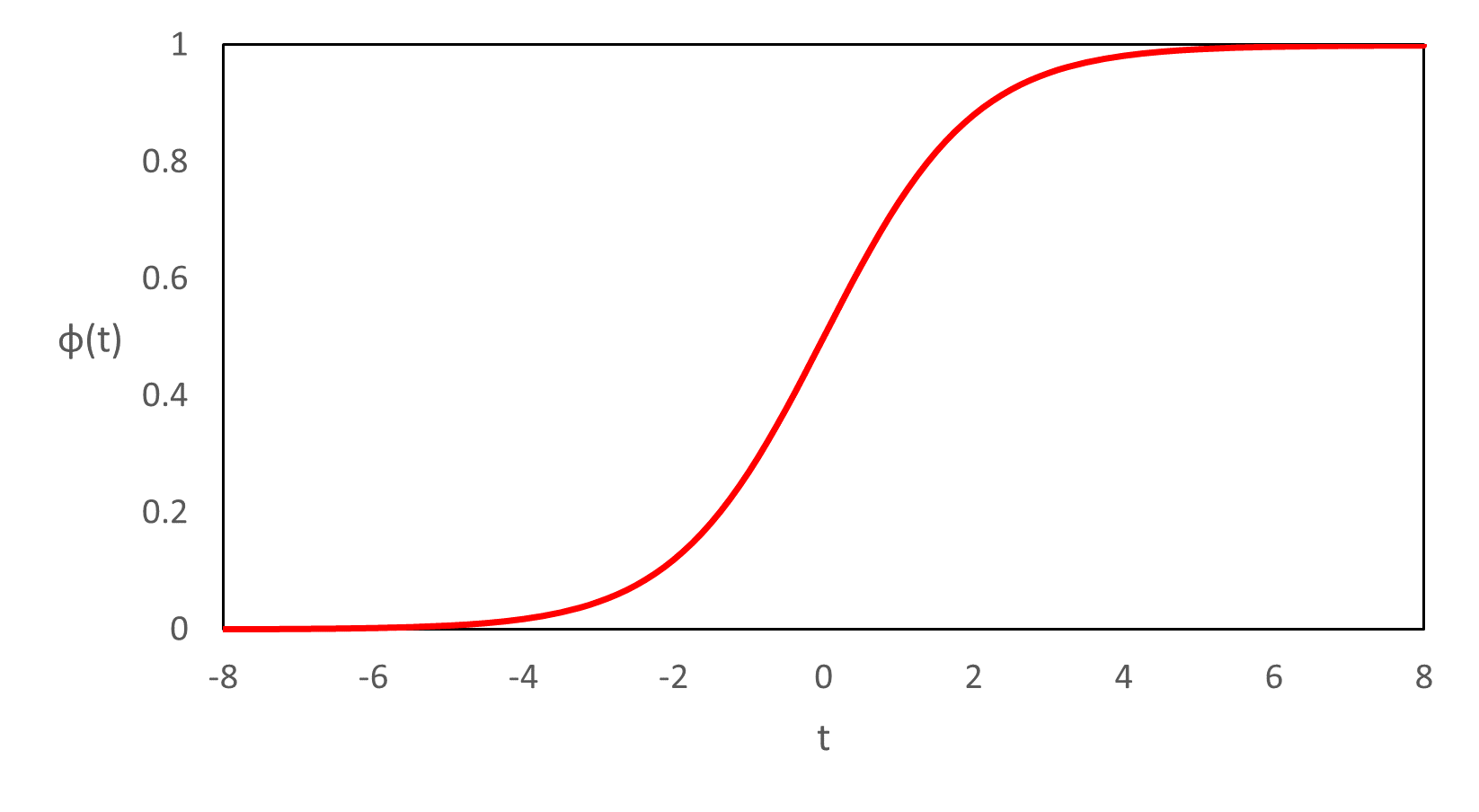}
  \caption{Sigmoid Activation Function of a Logistic Regression Model}
  \label{sigmoid}
\end{figure}

Due to the nature of our input and output data, we need a binary classifier and this is convenient as the logistic regression ML model uses a sigmoid normalizing function $\phi(t)$, 
\begin{equation}
\phi(t) = \frac{1}{1 + e^{-t}},
\end{equation}
\noindent as its activation function (visualized in Fig.~\ref{sigmoid}). The sigmoid function acts as a binary classifier.

Altogether, we obtain the net activation function for our model, $\phi(f(x, w))$, by composing Eqs. (1) and (2) as
\begin{equation}
\phi(f(x, w)) = \frac{1}{1 + e^{-\sum_{i=1}^{m} x_i w_i}}.
\end{equation}

Given our input layer, consisting of $x_i$ through $x_m$, and our weight layer of $w_i$ through $w_m$, our model can now feed them into one single neuron that uses the activation function to produce an output layer containing the vector $\hat{y}$. This concludes our forward pass in the model.

Rumelhart et al. [7] popularized the process of back-propagating errors in 1989, which repeatedly adjusts the weights in a model to minimize the error defined as the difference in the calculated output $\hat{y}$ and the expected output $y$. After each forward pass of the model as described above, backpropagation performs a backward pass and adjusts the weights in the weight matrix $W$ to reduce the error of the output.

We break down this process for logistic regression with an example of backpropagation.

The error vector after each forward pass is given by 
\begin{equation}
r = y - \hat{y}
\end{equation}
The derivative of the logistic regression model activation function is used to adjust the weights. This process is known as the error weighted derivative. The derivative of the sigmoid activation function can be simplified to
\begin{flalign}\nonumber
\frac{d(\phi(t))}{dt} & =\frac{0(1 + e^{-t})-1(-1(e^{-t}))}{(1 + e^{-t})^2}&\\ \nonumber
                      & =\frac{e^{-t}}{(1 + e^{-t})^{2}} =\frac{1 + e^{-t}}{(1 + e^{-t})^{2}} - \frac{1}{(1 + e^{-t})^{2}}&\\
                      & =\frac{1}{1 + e^{-t}}\left(1 - \frac{1}{1 + e^{-t}}\right) = \phi(t)\big(1 - \phi(t)\big)\nonumber
\end{flalign}
\begin{flalign}
\frac{d(\phi(t))}{dt} & =\phi(t)\big(1 - \phi(t)\big)&
\end{flalign}

Now we can create a matrix of adjustments $A$ which is equivalent to the product of the error vector $r$ and the sigmoid derivative of the output, $\phi(\hat{y})(1 - \phi(\hat{y}))$:
\begin{equation}
A = r \times \left[\phi(\hat{y})\big(1 - \phi(\hat{y})\big)\right]
\end{equation}

Finally, we update the weights by adding the dot product of the transpose of the input matrix $X^{T}$ and the adjustments matrix $A$ to our weight matrix $W$. This concludes our backward pass:
\begin{equation}
W = W + (X^{T} \cdot A)
\end{equation}

This process of forward and backward passes is repeated until we asymptotically reach a minimum error, and therefore a maximum accuracy.

\subsection{Model Tuning}

We proceeded to tune the model to improve accuracy. 
Tuning was done empirically by factoring in the time of day to choose the specific datapoints being used for training, as occupancy of stations during work hours differed from occupancy before and after work hours. Different time ranges of the dataset were used to train the model depending on the specific datapoint being tested. The pseudocode for how we changed the training dataset based on the time of day of a given test datapoint at a high-level is given in Algorithm 1.

\begin{algorithm}[hbt!]
\caption{Adjusting Training Datasets}\label{alg:cap}
\begin{algorithmic}
\If{$time < 0800$~hours}
    \State use $n_1$ data points starting from time $t_1$
    \State as training dataset
\ElsIf{$time < 1700$~hours}
    \State use $n_2$ data points starting from time $t_2$
    \State as training dataset
\Else
    \State use $n_3$ data points starting from time $t_3$
    \State as training dataset
\EndIf
\end{algorithmic}
\end{algorithm}



We needed a start-time to begin taking data points from our dataset for each prediction ($t_1$, $t_2$, and $t_3$ in Algorithm 1). The end-time for an input data range was the start-time plus the value of either $n_1$, $n_2$, or $n_3$, based on the time of the test datapoint. The start times were set to be $x - 23 - (h - v_i)$ for $i = 1, 2, 3$, where $h$ is the hour-of-day of the test datapoint and $v_i$ corresponded to the different threshold values of the work hours as defined earlier. Since a day starts at hour $0000$~hours, with work hours beginning at $0800$~hours and ending at $1700$~hours, $v_1 = 0$, $v_2 = 8$, and $v_3 = 17$. Therefore, the values of $t_1$, $t_2$, and $t_3$ were $x - 23 - h$, $x - 23 - (h - 8)$, and $x - 23 - (h - 17)$ respectively. For example, if we were to predict data early in the morning at $0400$~hours, data from 27 hours before ($0100$~hours on the previous day) was used, as $x - 23 - h = x - 23 - 4 = x - 27$.

The number of datapoints used for training in each of these sections were $n_1=10$, $n_2=12$, and $n_3=1$, respectively. These numbers were chosen as they resulted in the best model performance in comparison with all other combinations of 3 numbers chosen between 1 and 24 hours of the day. Note that the sector after $1700$~hours required significantly fewer datapoints compared to the other two sections, as occupancy of the charging stations was fairly stagnant after the end of the work day.


\subsection{Benchmark: Persistence Model}

As a benchmark model, we used a persistence model. A persistence model assumes that the future value of a time series is calculated under the assumption that nothing changes between the current time and the forecast time [8]. In our use-case, a persistence model assumed the occupancy of a charging station at a specific hour was the same as during the previous hour.

\subsection{Model Implementation}

For our ML pipeline, we used Scikit-learn, an open-source machine learning library for the Python programming language [9]. Scikit-learn further optimizes traditional loss and cost functions as well as the gradient descent method 
using Large-scale Bound-constrained Optimization and a Stochastic Average Gradient [10]-[12].

We include a sample schedule for scoring a logistic regression model with Scikit-learn in Listing 1. We omit data preparation and model tuning, among other parts of our ML pipeline, as we have already described these at a lower level. The sample schedule displays how the Sci-kit learn software was applied in this paper.

\definecolor{codegreen}{rgb}{0,0.6,0}
\definecolor{codegray}{rgb}{0.5,0.5,0.5}
\definecolor{codepurple}{rgb}{0.58,0,0.82}
\definecolor{backcolour}{rgb}{0.95,0.95,0.92}

\lstdefinestyle{mystyle}{
    backgroundcolor=\color{backcolour},   
    commentstyle=\color{codegreen},
    keywordstyle=\color{magenta},
    numberstyle=\tiny\color{codegray},
    stringstyle=\color{codepurple},
    basicstyle=\ttfamily\footnotesize,
    breakatwhitespace=false,         
    breaklines=true,                 
    captionpos=b,                    
    keepspaces=true,                 
    numbers=left,                    
    numbersep=5pt,                  
    showspaces=false,                
    showstringspaces=false,
    showtabs=false,                  
    tabsize=2
}

\lstset{style=mystyle}

\begin{lstlisting}[
    language=Python,
    caption=Sample Logistic Regression Schedule,
    mathescape,
    basicstyle=\small
]
# Define a logistic regression model using
# the Scikit-learn library
model = LogisticRegression()

# Train the model on the training datasets
model.fit(train_x, train_y)

# Predict the output for the test input
y_hat = model.predict(test_x)

# Evaluate the model score
score = accuracy_score(test_y, y_hat)
\end{lstlisting}

\section{RESULTS}

\subsection{Case Study}


For model training and testing, we gathered data from 57 ChargePoint EV charging stations from around the campus of University of California San Diego (UCSD) over 10 weeks, from the week of 6 January 2020 to the week of 9 March 2020. There were a total of 5,232 charging events and the average charging duration was 3 hours and 36 minutes. The overall percentage of time that the charging stations were occupied was 10.73\%. The data consisted of timestamps of vehicles plugging and unplugging at the different EV charging stations. 
The timestamped event data was processed into binary occupancy values for every hour interval in the 10 week range of the data for all of the 57 EV charging stations. Hours during which an EV was plugged in -- even if only for parts of the hour -- were assigned a 1 (occupied), otherwise the hour was marked as unoccupied, or 0. 
For the NOLR and persistence models, we used data from the week of 3 February 2020 to the week of 9 March 2020 for testing. In the NOLR case, training data was dynamically retrieved for each test datapoint from the 6-week test dataset as per Algorithm 1. For the persistence model, the training datapoint for each prediction was simply the previous datapoint in the test dataset, as explained in Section II.F. For traditional logistic regression, we used data from the week of 6 January 2020 to the week of 27 January 2020 for training, and data from the week of 3 February 2020 to the week of 9 March 2020 for testing.

\subsection{Model Scores}

\begin{figure}[thbp]
  \centering
  \includegraphics[scale=0.75]{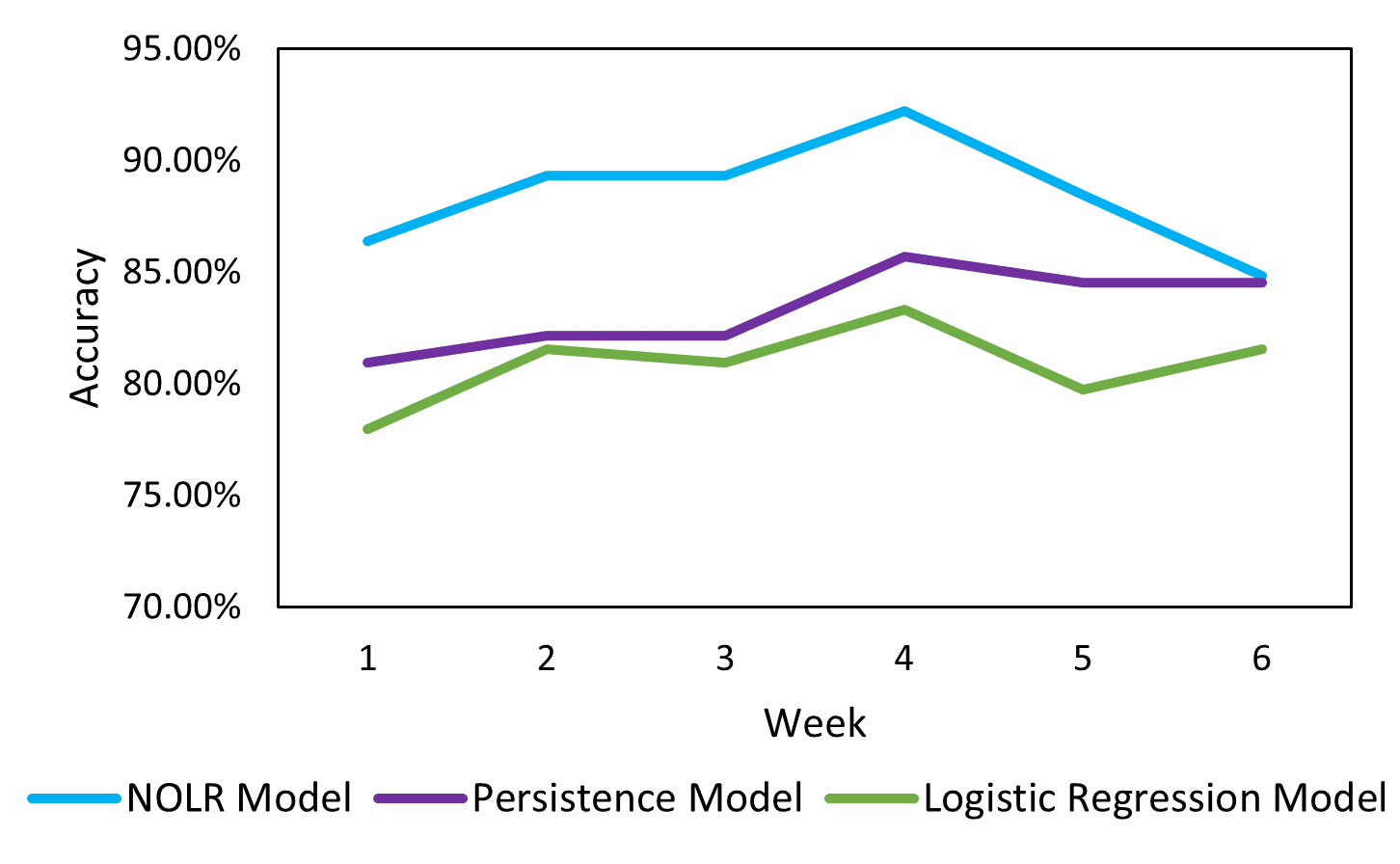}
  \caption{Comparison of Neighbor-Based Optimized Logistic Regression (NOLR) Model, Persistence Model, and Logistic Regression Model Accuracy Scores for EV Charging Station Occupancy Detection at UCSD}
  \label{chart}
\end{figure}

We compared the average accuracy scores of our NOLR model to the persistence and traditional logistic regression models, predicting the occupancy of one of the charging stations based on the 56 other surrounding stations, over ten 1-week ranges as plotted in Fig.~\ref{chart}. The NOLR model outperformed the other models on all accounts, attaining an average accuracy score of 88.43\% over the six testing weeks, compared to the persistence model's average accuracy of 83.33\% and the logistic regression model's average accuracy of 80.85\%. All three models reached their peak testing accuracy during week 4, where once again the NOLR model had the highest accuracy score of 92.23\%, versus the persistence model's 85.71\% and the logistic regression model's 83.33\%. Note that in all 6 testing weeks, the persistence model scores higher in accuracy than the regular logistic regression model. This is due to the fact that our data follows the same trends in each section of the work day (before 0800 hours, between 0800 hours and 1700 hours, and after 1700 hours) as described in the Model Tuning subsection. Therefore, since the persistence model uses the value of the datapoint before the test datapoint for its occupancy prediction, which is usually in the same section of the work day as the test datapoint, it results in accurate predictions. Exceptions to this occur closer to the edge cases (0800 hours, 1700 hours), when a persistence model prediction may use a datapoint from a different section of the work day.

\section{CONCLUSIONS}

\subsection{Final Remarks}

We draw three preeminent conclusions from the works of this paper. First, it is possible to predict the occupancy of an EV charging station not only based on its historic data, but also based on the occupancy of neighboring stations in real-time. The value of using neighbor-based occupancy prediction over occupancy prediction with historic data can be observed from the scoring difference between the 81.87\% accuracy value of the LSTM model by Ma et al. for predictions made 1 hour ahead, with our NOLR model average and maximum accuracy scores of 88.43\% and 92.23\% respectively. This highlights the fact that occupancy detection with neighbor-based optimized logistic regression is possible with higher accuracy than a LSTM occupancy detection model.

Second of the three principal conclusions lies in the architecture of the NOLR model. Rather than requiring a variety of layers with several neurons to process sequences of recurrent data, as is the case for LSTM, we simply needed a single-layer perceptron model to perform logistic regression, paired with an input data tuning algorithm based on the time of day, to produce our model. The NOLR model also had an average runtime of 0.65 seconds. As a result, our goal of building a streamlined model that only uses a logistical regression classifier while performing powerful predictions was also attained.

The third conclusion is a product of the former two, and elaborates on the importance of our work by relating it to energy conservation. 
The United States Environmental Protection Agency (US EPA) noted 
that an EV charger remains in standby mode for 85\% or more of its lifetime [13]. Standby power consumption can be minimized through occupancy prediction by selectively turning on and off charging stations that are in standby mode based on their predicted occupancy. 

\subsection{Future Work}

For the purpose of proving the viability of the model before implementing cross-platform capabilities, our model was trained and tested specifically for occupancy detection of EV charging stations. Neighbor-based optimized occupancy detection may be applied to other use cases such as predicting usage of plug loads to reduce standby power consumption from different appliances. Given the recent increase in decentralization of the electricity market and Distributed Energy Resources (DER), we can utilize our model for a variety of applications, including human occupancy detection in workplaces, schools, and households. 



\addtolength{\textheight}{-12cm}   










\end{document}